\documentclass[journal,article,submit,moreauthors]{Definitions/mdpi} 
\firstpage{1} 
\pubvolume{1}
\issuenum{1}
\articlenumber{0}
\pubyear{2026}
\copyrightyear{2026}
\datereceived{ } 
\daterevised{ } 
\dateaccepted{ } 
\datepublished{ } 

\usepackage{lineno}

\Title{PILOT: A Data-Free Continual Learning Approach for Real-Time Semantic Segmentation via Boundary Guidance}

\Author{Yujing Zhou $^{1,}$*, Prashant Shekhar $^{1}$, Thomas Yang $^{2}$ and Yongxin Liu $^{1}$}

\AuthorNames{Yujing Zhou, Prashant Shekhar, Thomas Yang and Yongxin Liu}

\address{%
$^{1}$ \quad Department of Mathematics, College of Arts and Sciences, Embry-Riddle Aeronautical University, Daytona Beach, FL 32114, USA; zhouy9@my.erau.edu (Y.Z.); shekharp@erau.edu (P.S.); liuy11@erau.edu (Y.L.)\\
$^{2}$ \quad Department of Electrical Engineering and Computer Science, College of Engineering, Embry-Riddle Aeronautical University, Daytona Beach, FL 32114, USA; yang482@erau.edu}

\corres{Correspondence: zhouy9@my.erau.edu}

\abstract{
Real-time semantic segmentation models offer an excellent balance between accuracy and inference speed. However, deploying these models in dynamic real world environments often requires the ability to learn novel classes incrementally without retraining on the entire dataset. This capability is known as continual learning. In this regard, the standard fine-tuning methods in deep learning often fail due to catastrophic forgetting, where the model learns new information but forgets previously trained and learned classes. Contributing to this crucial domain, the current paper proposes a novel continual learning framework tailored for PIDNet, which is a widely cited state-of-the-art real-time semantic segmentation model. Our method, PILOT(Parallel Incremental Learning Over Time), introduces a real-time and lightweight strategy by implementing a parallel Derivative-branch (D-branch) designed to capture the high frequency boundary information of novel classes while freezing the trained parameters of the original segmentation network. This novel setup allows the model to adapt to new semantic categories while preserving the knowledge of previously learned classes. By using only data associated with the new class, our model significantly reduces training overhead. Experimental results demonstrate that our approach successfully segments new classes while maintaining high mean Intersection over Union (mIoU) on the original base classes, thereby comfortably outperforming all major continual learning approaches in this domain. Overall, PILOT is shown to effectively mitigate catastrophic forgetting with minimal impact on inference latency, thus maintaining real-time performance.
}

\usepackage{algorithm}
\usepackage{algorithmic}
\begin{document}


\section{Introduction}

Semantic segmentation is a fundamental and important task in computer vision that divides an image into regions, where each pixel is assigned a semantic label (like "road," "car," or "pedestrian") and a corresponding mask is generated to isolate each identified object or area \cite{Long2015FCN}. It serves as a critical perception component for autonomous systems, including self-driving vehicles and mobile robots, which need to understand their surroundings in real time \cite{Cordts2016Cityscapes}. As these systems are deployed in diverse and evolving environments, the set of objects they must recognize unavoidably expands \cite{Parisi2019Lifelong}. For example, an autonomous driving vehicle trained to recognize sidewalks and pedestrians might later need to identify a new type of electric scooter or a temporary construction barrier when it enters another part of the city. Retraining the entire segmentation model from scratch every time a new class appears is time consuming, computationally expensive, and impractical for deployed systems \cite{Rebuffi2017iCaRL}. Therefore, the ability to learn new classes incrementally is essential for the stable and long term deployment of intelligent autonomous systems.

\begin{figure}[H]
    \centering
    \setlength{\fboxsep}{0pt} 
    \setlength{\fboxrule}{0.5pt} 
    
    \begin{minipage}{0.48\textwidth}
        \centering
        \fbox{\includegraphics[width=\linewidth, height=3cm]{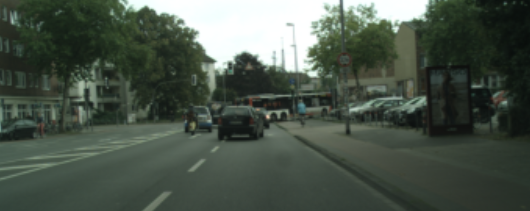}}
        \vspace{0.1cm}
        \centerline{(a) Original Image}
    \end{minipage}\hfill
    \begin{minipage}{0.48\textwidth}
        \centering
        \fbox{\includegraphics[width=\linewidth, height=3cm]{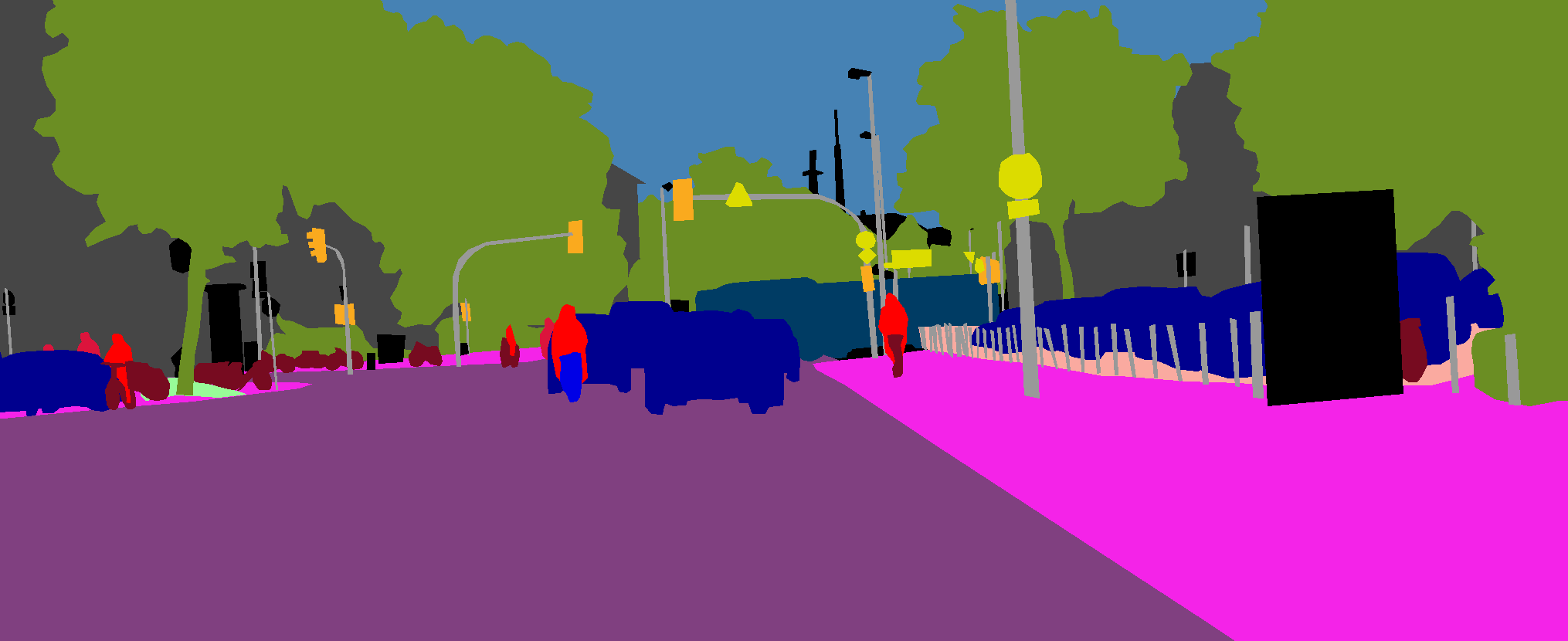}}
        \vspace{0.1cm}
        \centerline{(b) Ground Truth}
    \end{minipage}
    
    \vspace{0.4cm} 
    
    \begin{minipage}{0.48\textwidth}
        \centering
        \fbox{\includegraphics[width=\linewidth, height=3cm]{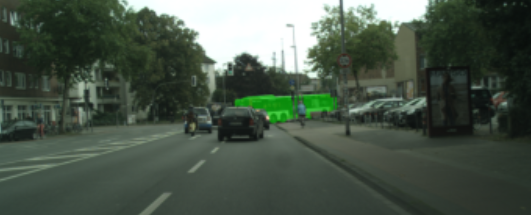}}
        \vspace{0.1cm}
        \centerline{(c) Proposed Method}
    \end{minipage}\hfill
    \begin{minipage}{0.48\textwidth}
        \centering
        \fbox{\includegraphics[width=\linewidth, height=3cm]{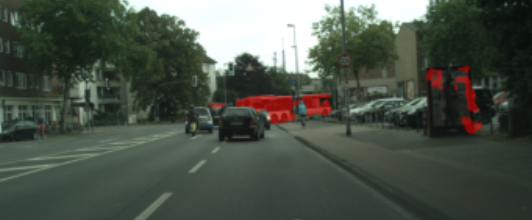}}
        \vspace{0.1cm}
        \centerline{(d) Catastrophic Forgetting}
    \end{minipage}
    
    \caption{Visual comparison of catastrophic forgetting versus the proposed continual learning framework. (a) The original input image. (b) The ground truth segmentation map. (c) The proposed method accurately segments the newly introduced ``Bus'' class (highlighted in green) without disrupting the scene. (d) Naive fine-tuning results in severe catastrophic forgetting, producing noisy and inaccurate predictions (highlighted in red) that confuse the new class with background elements like the billboard.}
    \label{fig:intro_forgetting}
\end{figure}

However, standard deep learning models like Convolutional Neural Networks (CNNs) usually lack the inherent ability to learn continually. When a model trained on a specific set of classes is updated with data from a new category, it tends to overwrite its existing parameters to minimize the loss on the new data. This problem is known as catastrophic forgetting in semantic segmentation \cite{McCloskey1989Catastrophic, French1999Catastrophic}, and it will result in a drastic performance drop on previously learned tasks. To mitigate this, researchers have proposed various Continual Learning (CL) strategies, ranging from data replay methods that store subsets of old data \cite{Rebuffi2017iCaRL} to regularization techniques that constrain weight updates \cite{Kirkpatrick2017EWC, Li2017LwF}. While effective in some scenarios, these methods often introduce significant memory overhead or fail to completely prevent forgetting in dense prediction tasks like semantic segmentation \cite{MiB}. Moreover, many existing CL approaches are designed for heavy, offline models and do not work well with lightweight, real-time models like our backbone model PIDNet \cite{Tan2025PIDNet} for this research, where model size and frames per second (FPS) are the most important considerations.

To explicitly illustrate the severity of catastrophic forgetting and the efficacy of the proposed solution, a direct visual comparison is presented in Figure \ref{fig:intro_forgetting}. When a standard semantic segmentation model is naively fine-tuned to learn a novel category—such as the ``Bus'' class—it suffers from severe parameter interference. As shown in Figure \ref{fig:intro_forgetting}(d), the naive approach completely loses its contextual understanding of the scene, producing noisy predictions (highlighted in red) that erroneously classify background elements, such as the billboard on the right, as the new object. In stark contrast, Figure \ref{fig:intro_forgetting}(c) demonstrates the output of the proposed continual learning framework. By structurally isolating the learning process, the model successfully identifies the new class with sharp boundaries (highlighted in green), matching the ground truth in Figure \ref{fig:intro_forgetting}(b) without corrupting the original spatial representations of the input image (Figure \ref{fig:intro_forgetting}(a)).

Therefore, to bridge the gap between heavy, offline continual learning methods and real-time segmentation models, a lightweight, real-time strategy for the state-of-the-art segmentation architecture is proposed. This approach utilizes the model's parallel branching structure to introduce plasticity for new classes while preserving stability for trained classes. The main contributions are summarized as follows:
\begin{itemize}
    \item A novel, replay-free continual learning framework for PIDNet is proposed that effectively mitigates catastrophic forgetting.
    \item It is demonstrated that the high-frequency boundary information captured by the D-branch is sufficient for learning new semantic categories without retraining the entire model.
    \item The proposed method (PILOT) is evaluated using a class-incremental learning scenario on the Cityscapes dataset, where the model first learns a set of base classes and subsequently adapts to new categories.
\end{itemize}

The remainder of this paper is organized as follows. Section \ref{sec:Related_Work} reviews related work in real-time semantic segmentation and continual learning. Section \ref{sec:Materials_and Methods} details the proposed continual learning architecture and training methodology. Section \ref{sec:Results} presents the experimental setup, datasets, and quantitative results. Finally, Section \ref{sec:Discussion} and \ref{sec:Conclusion} concludes the paper and discusses future directions.

\section{Related Work} \label{sec:Related_Work}
Real-time semantic segmentation has received considerable attention due to its importance in autonomous driving, robotics, and embedded perception. Early lightweight models such as ENet \cite{ENet} and ICNet \cite{ICNet} demonstrated that aggressive model compression and reduced-resolution processing can achieve high frame rates, although often with significant accuracy degradation. Subsequent approaches focused on multi-branch network designs to balance semantic context and spatial precision. For example, BiSeNet \cite{BiSeNet} introduced a dual-path architecture separating spatial detail and semantic reasoning. Expanding on this efficiency, models like Fast-SCNN \cite{FastSCNN} utilized shared shallow networks to reduce computational overhead, while STDC \cite{STDC} and DDRNet \cite{DDRNet} optimized spatial details and deep contextual features without relying on complex attention modules. More recently, PIDNet \cite{PIDNet} proposed a three-branch structure consisting of a P-branch for semantic context, an I-branch for intermediate feature refinement, and a D-branch dedicated to high-frequency boundary detection. This architecture provides state-of-the-art accuracy among real-time models, making it attractive for deployment in dynamic environments. The continued relevance of this architecture was recently highlighted by Tan and Jin \cite{Tan2025PIDNet}, who demonstrated that modifying PIDNet's proportional and integral branches with complex multi-scale feature fusion can further enhance baseline road sensing capabilities. These efforts primarily focus on scaling the contextual branches to maximize static, offline accuracy. Consequently, the potential of the lightweight D-branch for real-time continual adaptation remains largely unexplored. Indeed, similar to other standard real-time segmentation networks, the baseline PIDNet remains strictly trained offline and lacks inherent mechanisms for continual adaptation once deployed.

Beyond balancing speed and accuracy, accurately delineating object boundaries has emerged as a critical focus in modern segmentation architectures. Frameworks such as GSCNN \cite{GSCNN} introduced a dedicated shape stream to process spatial boundaries independently from the regular texture stream, proving that decoupling edge information significantly improves pixel-level classification along object borders. Similarly, PointRend \cite{PointRend} treated image segmentation as a rendering problem, selectively performing point-based predictions at ambiguous object boundaries to achieve crisp contours. These findings demonstrate that boundary-level features contain highly discriminative spatial cues. This fundamental characteristic is what makes the boundary-focused D-branch of PIDNet an ideal, lightweight anchor for learning new semantic categories.

Continual learning semantic segmentation aims to address the static nature of these deployed models by enabling them to learn new semantic categories over time without retraining from scratch or requiring access to previous datasets. This adaptability is especially critical because deployed systems frequently encounter highly dynamic and unstructured real-world environments where moving objects or novel obstacles can severely degrade the perception system's robustness, as observed in recent stereo vision frameworks \cite{Ai2023StereoSLAM}. A key challenge is catastrophic forgetting: when new-class data is introduced, old classes often appear as background, causing the model to overwrite their learned representations. Early pioneering works in this domain, such as ILT \cite{ILT}, attempted to adapt standard continual learning techniques by applying knowledge distillation to both the final output probabilities and the intermediate feature maps. Building on this, several frameworks have been proposed to specifically handle the background shift problem. MiB \cite{MiB} modeled the background shift during incremental steps by modifying the cross-entropy and distillation losses to better handle missing labels for old classes. PLOP \cite{PLOP} extended this idea using pooled output distillation, which aligns multi-scale intermediate features between teacher and student models to better preserve old-class knowledge.

Another direction incorporates pseudo-labeling and unknown-class modeling. SSUL \cite{SSUL} introduced an “unknown” label to identify unlabeled old-class regions and employed saliency-based masks to avoid collapsing old objects into background. IDEC \cite{IDEC} further advanced pseudo-labeling through dynamic thresholding and integrated adversarial and contrastive distillation (DADA and ARCL) to stabilize feature representations across incremental steps. These methods offer strong performance but generally rely on heavy backbones such as ResNet-101 \cite{ResNet} or DeepLab-style decoders \cite{DeepLab}, making them unsuitable for real-time settings.

Architectural approaches provide an alternative path. RCIL \cite{RCIL} introduced Representation Compensation Networks, where each convolutional block is decomposed into a frozen branch preserving old representations and a trainable branch responsible for learning new categories. Distillation is performed through pooled-cube features of the frozen teacher network. A related method, UCD \cite{UCD}, incorporated uncertainty-aware contrastive distillation to emphasize reliable feature regions during knowledge transfer. While effective, these frameworks also introduce computational overhead that makes them difficult to deploy in high-FPS applications.

Despite the progress of these continual learning semantic segmentation methods, two limitations persist: most are designed around high-capacity models unsuitable for embedded deployment, and their stability plasticity mechanisms do not translate well to lightweight architectures that prioritize speed. Many incorporate multi-scale feature distillation, auxiliary networks, or additional classifier heads that increase computational cost.

In contrast, this work focuses on continual learning for real-time segmentation by exploiting the inherent structure of PIDNet. Rather than adding external modules or heavy distillation pipelines, we leverage the model’s dedicated boundary focused D-branch to isolate plasticity while preserving stability in the P and I-branches. This enables the network to learn new semantic categories with minimal overhead and without relying on replay buffers or offline retraining. The proposed approach thus provides a lightweight and deployment friendly continual learning strategy for real-time models, addressing an important gap left by prior continual learning semantic segmentation frameworks.

\section{Materials and Methods} \label{sec:Materials_and Methods}

\subsection{Problem Formulation and Architecture Overview}

\begin{figure}[H]
\begin{adjustwidth}{-\extralength}{0cm}
\centering
\includegraphics[width=\linewidth, height=7cm]{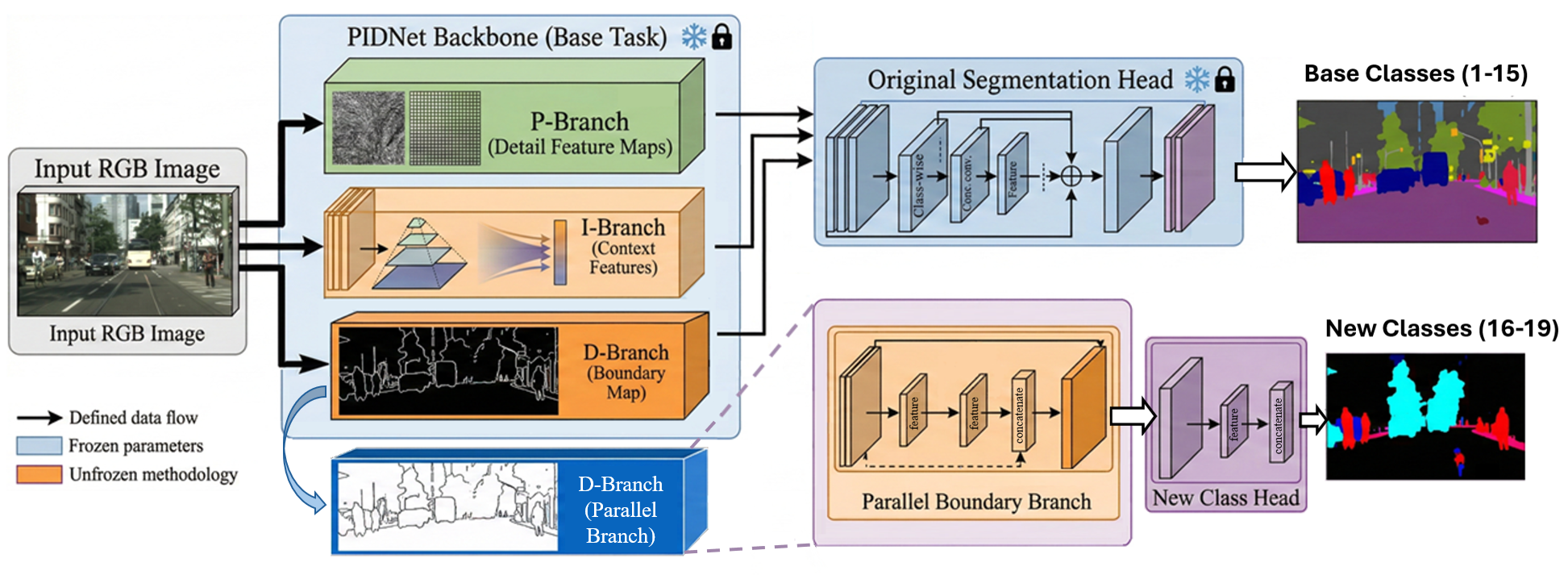}
\end{adjustwidth}
\caption{Overview of the proposed continual learning framework. The original PIDNet backbone and segmentation head (blue panels) are frozen after the base task to prevent catastrophic forgetting. To learn new semantic categories incrementally, an unfrozen Parallel Boundary Branch (purple panel) is introduced. This branch runs in parallel with the frozen D-Branch, sharing the same input from the RGB image, and bypasses the need to retrain the contextual P and I branches.\label{fig:architecture}}
\end{figure}

In class-incremental semantic segmentation, a model is trained on a sequence of tasks where each new task introduces novel semantic categories. Let $\mathcal{C}_0$ represent the set of base classes (e.g., the original 15 classes) learned during the initial training phase using dataset $\mathcal{D}_0$. In subsequent incremental steps $t \in \{1, \dots, T\}$, the model receives a new dataset $\mathcal{D}_t$ containing images annotated only for a new set of classes $\mathcal{C}_t$. The objective of continual learning is to update the model to accurately segment all encountered classes $\mathcal{C}_{0:t} = \mathcal{C}_0 \cup \dots \cup \mathcal{C}_t$ while strictly utilizing only the current dataset $\mathcal{D}_t$, without retraining on previous data. The primary challenge in this setting is preventing the catastrophic forgetting of earlier classes $\mathcal{C}_{0:t-1}$ while maintaining the plasticity to learn $\mathcal{C}_t$.

To address this, a lightweight, parallel-branch architecture built upon the PIDNet backbone is proposed. An overview of the PILOT framework is illustrated in Figure \ref{fig:architecture}. The architecture is conceptually divided into two distinct parts: a frozen base network and an unfrozen incremental framework.

The base network consists of the original PIDNet architecture, which includes the Proportional (P), Integral (I), and Derivative (D) branches, along with the original segmentation head. After the base training phase on $\mathcal{C}_0$, all parameters within this main backbone and original head are strictly frozen. This ensures that the foundational feature extraction capabilities and the semantic boundaries of the original 15 classes are permanently preserved, establishing a hard defense against catastrophic forgetting.

To accommodate novel classes, the Parallel Boundary Branch is introduced. As shown in Figure~\ref{fig:architecture}, this unfrozen branch runs in parallel with the frozen D-Branch, sharing the same input from the RGB image. By independently learning high-frequency boundary representations tailored to the novel classes, the parallel network can isolate and define the contours of new objects without interfering with the primary contextual pathways (P and I branches). The features from this parallel branch are then passed to a dedicated New Class Head, which outputs the segmentation masks for the incremental categories.

This dual-pathway design allows the model to route base class predictions through the original, undisturbed network while dynamically learning new categories through the isolated parallel structure.

\subsection{The Frozen PIDNet Backbone}
The foundation of the proposed framework relies on the PIDNet architecture, a real-time semantic segmentation model inspired by Proportional-Integral-Derivative (PID) controllers. To efficiently balance spatial details and global context, the network utilizes a three-branch structure:
\begin{itemize}
    \item \textbf{Proportional (P) Branch:} Designed to process and maintain semantic context across the image.
    \item \textbf{Integral (I) Branch:} Responsible for intermediate feature refinement, aggregating contextual information to support robust region classification.
    \item \textbf{Derivative (D) Branch:} Dedicated exclusively to high-frequency boundary detection, extracting the sharp edges and contours of semantic objects.
\end{itemize}

In a standard deployment, the features from these three branches are fused in the final stages to produce the segmentation map. However, in the proposed continual learning framework, the entire pre-trained PIDNet backbone and its original segmentation head are strictly frozen following the initial training phase on the base classes.

Freezing the base architecture serves two critical functions. First, it guarantees that the established feature representations and boundary definitions for the original classes remain entirely uncorrupted when new data is introduced. By preventing any weight updates in the main pathways, the model is immunized against catastrophic forgetting. Second, this constraint significantly reduces the computational overhead during incremental learning steps. Since gradients are not computed or updated for the massive base network, the training process remains lightweight. 

\subsection{The Parallel Boundary Branch}
To introduce plasticity for novel semantic categories without disrupting the frozen base network, a lightweight Parallel Boundary Branch is proposed. As established in the ablation study (Section 4.1), structurally aligning this incremental branch with the Derivative (D) branch of PIDNet yields the optimal balance between preserving base-class stability and learning new features.

The fundamental premise of this design is that high-frequency boundary information is particularly effective for delineating new semantic objects. Let $I \in \mathbb{R}^{3 \times H \times W}$ denote the input RGB image. In our framework, $I$ is processed by two structurally identical D-branches that operate in parallel: the original frozen D-branch produces a boundary feature map $F_D \in \mathbb{R}^{C \times H \times W}$ used by the original segmentation head for base-class prediction, while an unfrozen parallel D-branch independently produces a novel-class boundary feature map $F_{D'} \in \mathbb{R}^{C \times H \times W}$ from the same input. This parallel design ensures that the boundary representations learned for novel classes do not constrain or modify the frozen base-class features.

The parallel branch shares the architectural design of the original D-branch but is initialized as an unfrozen module that learns class-specific boundary representations for novel categories. As depicted in the architectural diagram, its internal structure consists of sequential feature extraction blocks and a concatenation layer, which aggregate multi-scale boundary details. Mathematically, this feature transformation can be expressed as:
$$ F_{D'} = \Phi_{parallel}(I; \theta_{parallel}) $$
where $\Phi_{parallel}$ represents the operations within the parallel branch and $\theta_{parallel}$ denotes its trainable parameters.

The novel-class boundary feature map, $F_{D'}$, is then fed into a dedicated New Class Head. This independent segmentation head is specifically responsible for generating the final spatial logits, $Y_{new}$, for the newly introduced classes $\mathcal{C}_t$:
$$ Y_{new} = \mathcal{H}_{new}(F_{D'}; \theta_{head}) $$
where $\mathcal{H}_{new}$ and $\theta_{head}$ represent the new segmentation head and its respective parameters.

By isolating the learning process strictly within $\theta_{parallel}$ and $\theta_{head}$, the proposed method ensures that the gradient updates triggered by the novel class data never backpropagate into the base PIDNet backbone. This structural isolation effectively eliminates catastrophic forgetting while requiring only a fraction of the computational overhead compared to retraining the entire network or utilizing heavy distillation pipelines.

\subsection{Training Strategy and Loss Function}
During the incremental learning phase for a new task $t$, the model is optimized exclusively using the novel dataset $\mathcal{D}_t$. Unlike traditional continual learning methods that mitigate forgetting by maintaining memory-intensive replay buffers of historical data, the proposed framework is entirely replay-free.

\begin{algorithm}[H]
\caption{Incremental Training via the Parallel Boundary Branch}
\label{alg:incremental_training}
\begin{algorithmic}[1]
\REQUIRE Pre-trained PIDNet backbone $\mathcal{B}$ with parameters $\theta_{base}$
\REQUIRE Pre-trained original head $\mathcal{H}_{orig}$ with parameters $\theta_{orig\_head}$
\REQUIRE Incremental dataset $\mathcal{D}_t$ containing images $X$ and labels $Y$ for novel classes $\mathcal{C}_t$
\REQUIRE Learning rate $\alpha$, total epochs $E$

\STATE \textbf{Phase 1: Initialization}
\STATE Freeze backbone parameters: $\theta_{base} \leftarrow \text{requires\_grad}(False)$
\STATE Freeze original head: $\theta_{orig\_head} \leftarrow \text{requires\_grad}(False)$
\STATE Initialize Parallel Boundary Branch parameters: $\theta_{parallel}$
\STATE Initialize New Class Head parameters: $\theta_{head}$

\STATE \textbf{Phase 2: Incremental Optimization}
\FOR{$epoch = 1$ \TO $E$}
    \FOR{each mini-batch $(X, Y) \in \mathcal{D}_t$}
        \STATE \textit{1. Forward pass through unfrozen parallel branch}
        \STATE Generate novel-class boundary features: $F_{D'} \leftarrow \Phi_{parallel}(X; \theta_{parallel})$
        \STATE Generate novel class predictions: $\hat{Y}_{new} \leftarrow \mathcal{H}_{new}(F_{D'}; \theta_{head})$
        
        \STATE \textit{2. Loss computation (only on novel classes)}
        \STATE Compute Cross-Entropy Loss: $\mathcal{L}_{inc} = \text{CE}(\hat{Y}_{new}, Y \in \mathcal{C}_t)$
        
        \STATE \textit{3. Backpropagation and targeted parameter update}
        \STATE Compute gradients: $\nabla \theta_{parallel}, \nabla \theta_{head} \leftarrow \frac{\partial \mathcal{L}_{inc}}{\partial \theta_{parallel}}, \frac{\partial \mathcal{L}_{inc}}{\partial \theta_{head}}$
        \STATE Update parallel branch: $\theta_{parallel} \leftarrow \theta_{parallel} - \alpha \nabla \theta_{parallel}$
        \STATE Update new head: $\theta_{head} \leftarrow \theta_{head} - \alpha \nabla \theta_{head}$
    \ENDFOR
\ENDFOR
\RETURN Updated incremental parameters $\theta_{parallel}$ and $\theta_{head}$
\end{algorithmic}
\end{algorithm}

The complete step-by-step training protocol is summarized in Algorithm \ref{alg:incremental_training}. This protocol strictly limits weight updates to the newly introduced modules. During the initialization phase, all parameters within the foundational PIDNet backbone ($\theta_{base}$) and the original segmentation head ($\theta_{orig\_head}$) are permanently frozen. Consequently, during backpropagation, gradients are computed and applied solely to the parameters of the Parallel Boundary Branch ($\theta_{parallel}$) and the New Class Head ($\theta_{head}$). 

To optimize these incremental components, a standard pixel-wise Cross-Entropy (CE) loss is applied. Let $y_i \in \mathcal{C}_t$ denote the ground-truth label for pixel $i$ in a given image from the incremental dataset, and let $\hat{y}_{i,c}$ represent the predicted probability that pixel $i$ belongs to the novel class $c$, derived from the softmax activation of the New Class Head. The incremental loss function $\mathcal{L}_{inc}$ is mathematically formulated as:
$$ \mathcal{L}_{inc} = - \frac{1}{N} \sum_{i=1}^{N} \sum_{c \in \mathcal{C}_t} \mathbb{1}(y_i = c) \log(\hat{y}_{i,c}) $$
where $N$ represents the total number of valid pixels in the spatial dimensions, and $\mathbb{1}(\cdot)$ is the indicator function, which equals 1 if the condition is true and 0 otherwise. While the framework supports adding multiple novel classes per incremental step, our experimental setup adds one class at a time, i.e., $|\mathcal{C}_t| = 1$ for all $t$.

Because the base classes $\mathcal{C}_{0:t-1}$ are absent from the incremental dataset $\mathcal{D}_t$, the loss is computed strictly over the spatial regions annotated for the novel classes $\mathcal{C}_t$. By restricting the loss computation and weight updates exclusively to the lightweight parallel modules, the training process avoids the massive computational burden associated with full-network fine-tuning or complex multi-scale feature distillation pipelines. This targeted optimization ensures rapid, resource-efficient adaptation to novel semantic categories, maintaining the low inference latency required for real-time deployment.

\vspace{0.3cm}
\begin{figure}[H]
\begin{adjustwidth}{-\extralength}{0cm}
\centering
\includegraphics[width=\linewidth]{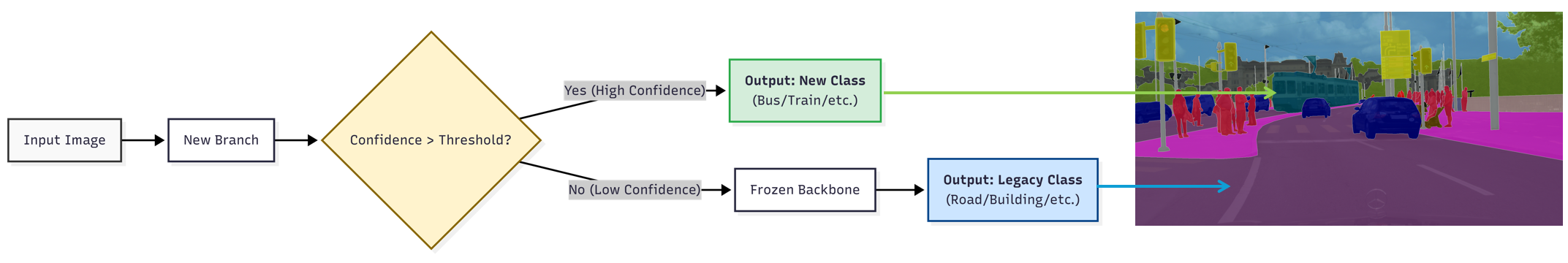} 
\end{adjustwidth}
\caption{The inference prediction routing mechanism. The model first evaluates the input image through the new parallel branch. Pixels exceeding the confidence threshold are classified as the new category. For low confidence regions, the model falls back to the frozen backbone to accurately predict legacy classes, effectively merging the two pathways into a single, coherent segmentation map.\label{fig:inference_flow}}
\end{figure}
\vspace{0.4cm}

To generate the final segmentation map during deployment, a confidence based routing mechanism is utilized to seamlessly merge the predictions from the unfrozen parallel branch and the frozen base network. The complete inference data flow is illustrated in Figure \ref{fig:inference_flow}. 

When an input image is processed, the parallel boundary branch first evaluates the scene to detect any newly learned semantic categories. The dedicated new class head outputs a probability map for these novel objects. A predetermined confidence threshold ($\tau$) is then applied to these predictions. If the confidence score for a specific pixel exceeds this threshold, the model assigns it to the novel class (e.g., Bus or Train). 

Conversely, if the confidence score falls below the threshold, the system defaults to the foundational knowledge of the frozen PIDNet backbone. The frozen network then processes those low confidence regions to assign legacy classes (e.g., Road or Building). This conditional routing strategy ensures that novel objects are only segmented when the model is highly certain. It strictly preserves the integrity of the base classes in all other areas. By intelligently routing the predictions rather than aggressively overwriting them, the framework maintains the high accuracy and real-time efficiency required for autonomous perception.

\section{Results} \label{sec:Results}
We evaluate the PILOT framework on the Cityscapes dataset~\cite{Cordts2016Cityscapes}, a large-scale benchmark for urban scene understanding. Cityscapes contains 5,000 finely annotated images collected from 50 cities across Germany and neighboring countries, split into 2,975 training images, 500 validation images, and 1,525 test images. Each image has a resolution of 2048 $\times$ 1024 pixels and is annotated with pixel-level labels across 19 semantic classes, including road, sidewalk, building, vegetation, person, car, bus, train, motorcycle, and bicycle, among others. Following standard continual semantic segmentation protocols, we use the 2,975 training images for both base training and incremental learning steps, and report all mIoU scores on the 500 validation images.

The following subsections represent a comprehensive evaluation of the PILOT framework on the Cityscapes dataset under a class-incremental semantic segmentation protocol. We begin in Section~\ref{sec:continual_learning} by analyzing the long-term continual learning behavior of PILOT across all three PIDNet variants (S, M, and L), comparing against Joint Training (upper bound) and Fine-Tuning (lower bound) baselines under a matched-backbone protocol. We then position PILOT against state-of-the-art continual semantic segmentation methods on the 14-1 and 10-1 benchmarks in Section~\ref{sec:comparison_sota}. Section~\ref{sec:dynamic_stability} examines the stability-plasticity trade-off through per-step mIoU trajectories. Section~\ref{sec:ablation} provides an ablation analysis of the proposed parallel branch connection point, and Section~\ref{sec:threshold} examines the sensitivity of the model's performance to the decision threshold of the new class head. Finally, Section~\ref{sec:qualitative} presents qualitative segmentation results visualizing the model's incremental learning behavior across model capacities.

\subsection{Continual Learning} \label{sec:continual_learning}
We evaluate the long-term continual learning capability of our method by extending the experiment from the base 15 classes to the full 19 classes, adding one class at a time across four incremental steps. The quantitative results at each incremental stage are detailed in Table~\ref{tab:incremental_results_15to19}. To contextualize the performance of our method, we establish reference baselines using Joint Training and Fine-Tuning, both implemented on the PIDNet-M backbone for fair comparison.

\begin{table}[H]
\caption{Quantitative analysis of incremental learning (Steps 15 to 19). \textbf{Part I} reports the Overall mIoU (all classes known at that step). \textbf{Part II} reports the Base Class Stability (mIoU calculated only on the original 15 classes), showing how well the model resists forgetting.\label{tab:incremental_results_15to19}}
	\begin{adjustwidth}{-\extralength}{0cm}
		\begin{tabularx}{\fulllength}{LCCCCCC}
			\toprule
			 & & \textbf{Base Task} & \multicolumn{4}{c}{\textbf{Incremental Steps}} \\
			\cmidrule(lr){3-3} \cmidrule(lr){4-7}
			\textbf{Method} & \textbf{Backbone} & \textbf{Step 0 (15 cls)} & \textbf{Step 1 (16 cls)} & \textbf{Step 2 (17 cls)} & \textbf{Step 3 (18 cls)} & \textbf{Step 4 (19 cls)} \\
			\midrule
			Joint Training & PIDNet-M & 80.92 & 78.90 & 78.50 & 78.10 & 77.80 \\
			Fine Tuning    & PIDNet-M & 80.92 & 70.08 & 65.12 & 61.75 & 59.90 \\
			\midrule
			\multicolumn{7}{c}{\textbf{Part I: Overall Performance (mIoU on All Classes)}} \\
			\midrule
			PILOT & PIDNet-S & 79.02 & 72.71 & 69.94 & 68.25 & 68.91 \\
			\textbf{PILOT} & \textbf{PIDNet-M} & \textbf{80.92} & \textbf{74.80} & \textbf{72.03} & \textbf{71.71} & \textbf{71.87} \\
			PILOT & PIDNet-L & 80.58 & 75.00 & 72.03 & 70.97 & 70.54 \\
			\midrule
			\multicolumn{7}{c}{\textbf{Part II: Base Class Stability (mIoU on Original 15 Classes Only)}} \\
			\midrule
			PILOT & PIDNet-S & 79.02 & 77.24 & 75.32 & 75.21 & 75.19  \\
			\textbf{PILOT} & \textbf{PIDNet-M} & \textbf{80.92} & \textbf{79.14} & \textbf{77.37} & \textbf{77.90} & \textbf{77.38} \\
			PILOT & PIDNet-L & 80.58 & 79.31 & 77.32 & 77.29 & 76.76 \\
			\bottomrule
		\end{tabularx}
	\end{adjustwidth}
\end{table}

\textbf{Joint Training (Upper Bound).} This setting represents the ideal offline scenario in which the model is retrained from scratch on the full accumulated dataset whenever a new class is introduced. While this yields strong performance (maintaining approximately 78\% mIoU throughout), it requires access to all historical data at every step and incurs substantial retraining cost. Joint Training is therefore impractical for real-world autonomous driving deployment and serves only as a theoretical upper bound.

\textbf{Fine-Tuning (Lower Bound).} This setting represents a naive incremental approach in which the model is sequentially fine-tuned on new class data without access to previous samples or knowledge-preservation mechanisms. As shown in Table~\ref{tab:incremental_results_15to19}, this approach suffers from progressive performance degradation, dropping from 80.92\% to 59.90\% over four incremental steps. This confirms that without an explicit mechanism for preserving prior knowledge, incremental training is inadequate for long-term class addition.

\textbf{Efficacy of Our Approach.} PILOT demonstrates robust stability across all sequential incremental steps. Utilizing the PIDNet-M backbone, which demonstrates the optimal balance between capacity and plasticity in our experiments, we observe a consistent performance trajectory. At the first incremental step (16 classes), the model achieves an overall mIoU of 74.80\%, effectively retaining base knowledge with a base-class stability of 79.14\%. This stability persists over the long term: even after four sequential additions (19 classes), the model maintains an overall mIoU of 71.87\%, with base-class mIoU at 77.38\%, a retention of 95.6\% relative to the initial 80.92\% base performance. This indicates that our D-branch connection successfully mitigates catastrophic forgetting while allowing the model to accommodate novel classes. Furthermore, the PIDNet-M variant yields superior performance compared to both the S and L backbones. At the final step, PIDNet-M outperforms the Fine-Tuning baseline by 11.97 percentage points.

\subsection{Comparison with State-of-the-Art Methods} \label{sec:comparison_sota}
We compare our method against representative continual semantic segmentation (CSS) methods on Cityscapes under two standard protocols: 14-1 (6 tasks) and 10-1 (10 tasks). Table~\ref{tab:cityscapes_css} reports the mean IoU over the initial (base) classes, the novel (incremental) classes, and all classes at the final incremental step.

Before discussing the results, we briefly explain the CSS evaluation protocol. The notation ``$B$-$N$'' denotes an experimental setup in which $B$(Base) classes are jointly learned during the base training phase and $N$(New) classes are added one at a time during incremental learning. Under the 14-1 protocol, the model is first trained on 14 base classes, after which the remaining 5 classes (15 through 19) are introduced sequentially across 5 incremental steps, yielding 6 total tasks. Similarly, the 10-1 protocol trains on 10 base classes and adds the remaining 9 classes one at a time, for 10 total tasks. All mIoU values reported in Table~\ref{tab:cityscapes_css} are measured on the \emph{final model} after all incremental steps are complete, but evaluated over different class subsets. Columns labeled ``1-14'' and ``1-10'' report the mIoU computed only over the base classes, measuring how well each method resists catastrophic forgetting. Columns labeled ``15-19'' and ``11-19'' report the mIoU computed only over the novel (incrementally added) classes, measuring how well each method learns new categories. The columns labeled ``all'' report the mIoU computed over all 19 classes, indicating overall segmentation performance.

\begin{table}[H]
\caption{Continual semantic segmentation results on Cityscapes in mIoU (\%). Tasks defined as $C^1$-$C^T$ ($T$ tasks).\label{tab:cityscapes_css}}
	\begin{adjustwidth}{-\extralength}{0cm}
		\begin{tabularx}{\fulllength}{LCCCCCC}
			\toprule
			 & \multicolumn{3}{c}{\textbf{14-1 (6 tasks)}} & \multicolumn{3}{c}{\textbf{10-1 (10 tasks)}} \\
			\cmidrule(lr){2-4} \cmidrule(lr){5-7}
			\textbf{Method} & \textbf{1-14} & \textbf{15-19} & \textbf{all} & \textbf{1-10} & \textbf{11-19} & \textbf{all} \\
			\midrule
			PLOP~\cite{PLOP}       & 63.54 & 15.38 & 48.33 & 60.75 & 27.97 & 42.96 \\
			MiB~\cite{MiB}         & 66.37 & 14.36 & 50.05 & 61.80 & 32.97 & 45.73 \\
			MiB + AWT~\cite{AWT}   & 65.60 & 19.19 & 50.72 & 60.97 & 35.70 & 46.55 \\
			DKD~\cite{DKD}         & 68.83 & 14.70 & 51.86 & 66.77 & 34.52 & 48.92 \\
			TOPICS~\cite{TOPICS}   & 73.03 & 42.47 & 61.74 & 71.37 & 52.62 & 59.36 \\
			\midrule
			PILOT (Ours) & \textbf{76.52} & \textbf{49.20} & \textbf{70.21} & \textbf{75.84} & \textbf{54.20} & \textbf{65.68} \\
			\bottomrule
		\end{tabularx}
	\end{adjustwidth}
\end{table}

It is important to note that prior CSS methods including PLOP~\cite{PLOP}, MiB~\cite{MiB}, MiB+AWT~\cite{AWT}, DKD~\cite{DKD}, and TOPICS~\cite{TOPICS} adopt DeepLabV3 with a ResNet-101 backbone, whereas our method uses PIDNet for its real-time efficiency in autonomous driving deployment. While this backbone difference precludes strictly equivalent comparison, two observations indicate the effectiveness of our continual learning framework. First, our method achieves substantially stronger base-class retention: on the 14-1 protocol, we obtain 76.52\% on classes 1-14 compared to TOPICS' 73.03\% and PLOP's 63.54\%. Second, our method demonstrates markedly improved plasticity on novel classes, achieving 49.20\% on classes 15-19, compared to 42.47\% for TOPICS, 19.19\% for MiB+AWT, and 15.38\% or lower for the remaining baselines. The gain is consistent across both the 14-1 and 10-1 protocols.

\begin{figure}[H]
\centering
\includegraphics[width=\linewidth]{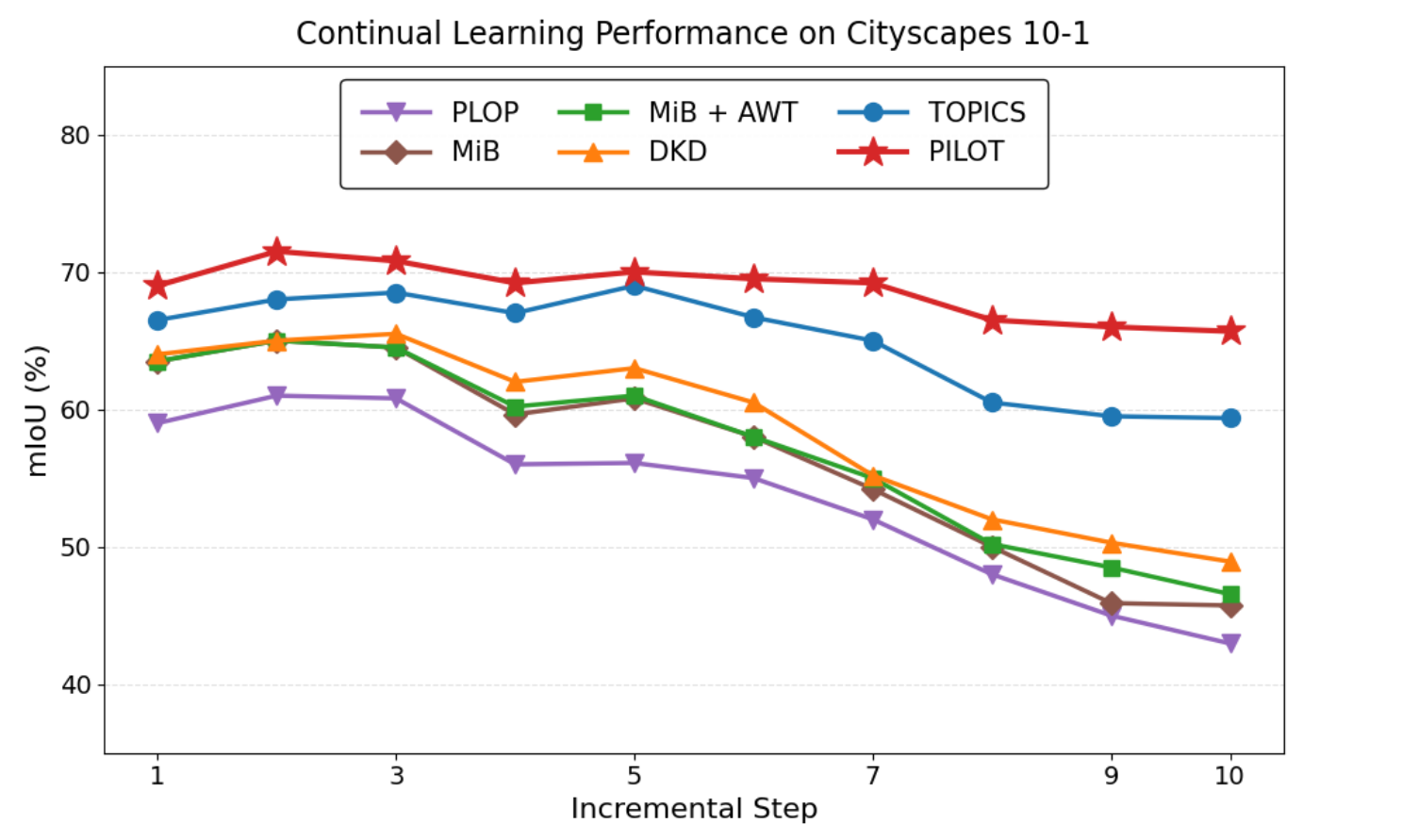}
\caption{Per-step overall mIoU on the Cityscapes 10-1 protocol across nine sequential incremental steps. Our method (red) maintains consistently higher and more stable performance than all prior CSS approaches throughout the entire incremental sequence. Baseline curves were digitized from Hindel et al.~\cite{TOPICS} (Figure 5).\label{fig:miou_comp}}
\end{figure}

To further visualize the dynamics of catastrophic forgetting across methods, we plot per-step overall mIoU on the 10-1 protocol in Figure~\ref{fig:miou_comp}. The 10-1 setting is particularly informative because it requires nine sequential incremental steps, providing a clearer view of long-term stability than shorter protocols. As shown in the figure, all baseline methods exhibit a pronounced downward trajectory as more classes are added, with PLOP, MiB, and MiB+AWT degrading sharply after Step 6 and ending below 50\% mIoU. TOPICS maintains the strongest performance among prior approaches, but still declines substantially in later steps, falling from approximately 69\% at Step 5 to 59.36\% at Step 9. In contrast, our method maintains consistently higher mIoU throughout the entire incremental sequence, ending at 65.68\% at Step 9, a margin of 6.32 percentage points over TOPICS. Notably, our curve also exhibits a flatter trajectory, indicating that the per-step performance loss from each incremental step is smaller than for the baselines. This stability is consistent with our base-class retention results in Table~\ref{tab:cityscapes_css} and supports our claim that the proposed framework more effectively mitigates catastrophic forgetting under long incremental sequences.

These results, together with the matched-backbone ablation in Table~\ref{tab:incremental_results_15to19}, suggest that our framework addresses both the stability-plasticity trade-off and the semantic background shift problem more effectively than prior approaches on Cityscapes continual segmentation benchmarks.

\subsection{Dynamic Stability Analysis} \label{sec:dynamic_stability}
\begin{figure}[H]
\includegraphics[width=\linewidth]{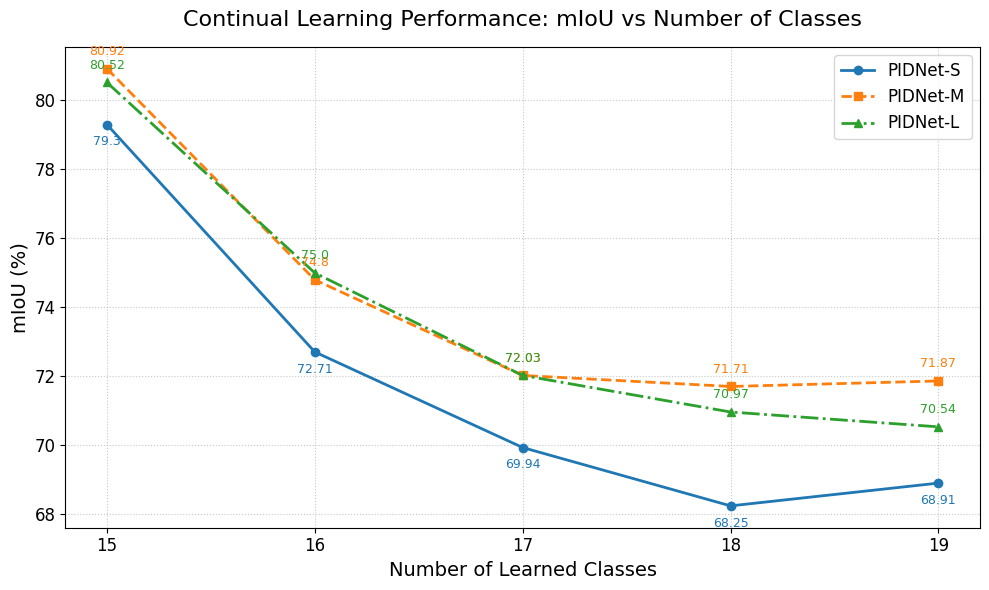} 
\caption{Continual Learning Performance (mIoU vs. Number of Classes). The plot illustrates the performance trajectory of PILOT with PIDNet-S, M, and L backbone variants across incremental steps, bounded by the Joint Training upper bound and Fine-Tuning lower bound. PIDNet-M (dashed line) demonstrates the highest robustness among our variants, stabilizing above 71\% at the final step.\label{fig:miou_curves}}
\end{figure}

To further investigate the stability-plasticity trade-off, we analyze the mIoU evolution across sequential incremental steps in Figure~\ref{fig:miou_curves}.

As shown in the plot, all three model variants experience a noticeable drop in performance when the first new class is introduced (from 15 to 16 classes). This initial decline is expected, as the model transitions from a closed-set training regime to incremental learning. Several factors contribute to this drop. First, the parallel boundary branch and new class head are randomly initialized and require training to produce reliable predictions for the novel class. Second, the confidence-based routing mechanism may occasionally misclassify base-class pixels as the novel category before the new head is fully calibrated. Third, the mIoU computation itself changes scale as additional classes are introduced into the evaluation. Notably, our framework largely avoids the classical semantic background shift problem that plagues unified-classifier CSS methods, since the frozen original head and the new class head operate on separate prediction pathways.

Critically, performance stabilizes significantly after this first step. From 17 classes onward, the decline becomes substantially slower and the curve flattens. This indicates that our D-branch connection effectively limits catastrophic forgetting once the model has adjusted to the incremental setting. By isolating boundary features, the network accommodates new classes without aggressively overwriting the feature representations of the original 15 categories.

Regarding model capacity, PIDNet-M consistently achieves higher mIoU than both PIDNet-S and PIDNet-L across incremental steps. We hypothesize that PIDNet-M's parameter count offers a favorable balance: sufficient capacity to learn discriminative features for new classes, while remaining constrained enough to resist excessive interference with previously learned representations. In contrast, the larger PIDNet-L may be more prone to overfitting on the limited novel-class data. However, the absolute differences among the three variants are small (PIDNet-M: 71.87\%, PIDNet-L: 70.54\%), suggesting that our framework is relatively robust across model capacities. A more detailed investigation of this relationship is left to future work.

\subsection{Ablation Analysis} \label{sec:ablation}
PIDNet~\cite{PIDNet} decomposes feature learning into three parallel branches: the Proportional (P) branch, which preserves high-resolution detail; the Integral (I) branch, which aggregates contextual information through low-resolution processing; and the Derivative (D) branch, which extracts high-frequency boundary features to guide segmentation. To determine the optimal injection point for the parallel incremental branch, we conducted an ablation study by attaching the branch to each of these three branches of the frozen PIDNet backbone. The quantitative results are summarized in Table~\ref{tab:branch_comparison}.

As shown in the table, the D-branch configuration yields the best overall performance, achieving a final mIoU of 72.65\% on the 16-class task. This significantly outperforms the P-branch (67.97\%) and the I-branch (70.25\%).

In terms of stability, all three connection points preserve base-class accuracy within a narrow range (P: 76.75\%, I: 77.49\%, D: 77.22\%), a modest decrease from the 79.02\% baseline. However, the D-branch substantially outperforms the other two variants in learning the new class, yielding the highest overall mIoU (72.65\%) compared to the I-branch (70.25\%) and P-branch (67.97\%). These results confirm our hypothesis that high-frequency boundary information is particularly valuable for delineating new objects in an incremental learning setting, making the D-branch the optimal connection point.

\begin{table}[H] 
\caption{Ablation Analysis of connection points. The D-Branch (highlighted) achieves the best trade-off between stability (Base mIoU) and plasticity (Overall mIoU).\label{tab:branch_comparison}}
\begin{tabularx}{\textwidth}{LCCC}
\toprule
 & \textbf{Baseline} & \multicolumn{2}{c}{\textbf{Incremental Step (After Adding Class)}} \\
\cmidrule(lr){3-4}
\textbf{Connection} & \textbf{Original mIoU} & \textbf{Base mIoU} & \textbf{Overall mIoU} \\
\textbf{Point} & \textbf{(15 Classes)} & \textbf{(Old Classes)} & \textbf{(All 16 Classes)} \\
\midrule
P Branch & 79.02\% & 76.75\% & 67.97\% \\
I Branch & 79.02\% & 77.49\% & 70.25\% \\
\textbf{D Branch} & \textbf{79.02\%} & \textbf{77.22\%} & \textbf{72.65\%} \\
\bottomrule
\end{tabularx}
\end{table}

\subsection{Hyperparameter Analysis} \label{sec:threshold}

To maximize the segmentation performance of the newly introduced classes, the decision threshold ($\tau$) for the Parallel Boundary Branch was systematically evaluated. Because the new class head outputs continuous probability maps, selecting an optimal $\tau$ is critical for balancing false positives and false negatives during inference.

\begin{figure}[H]
    \centering
    \includegraphics[width=0.8\linewidth, height=8cm]{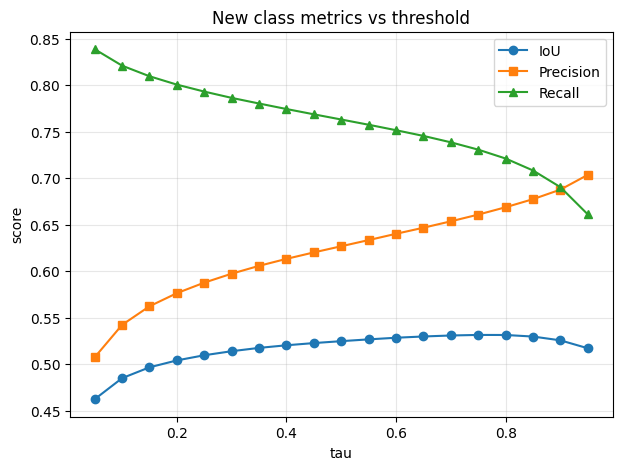}
    \\ \vspace{0.1cm} (a)
    
    \vspace{0.2cm}
    
    \includegraphics[width=0.7\linewidth, height=8cm]{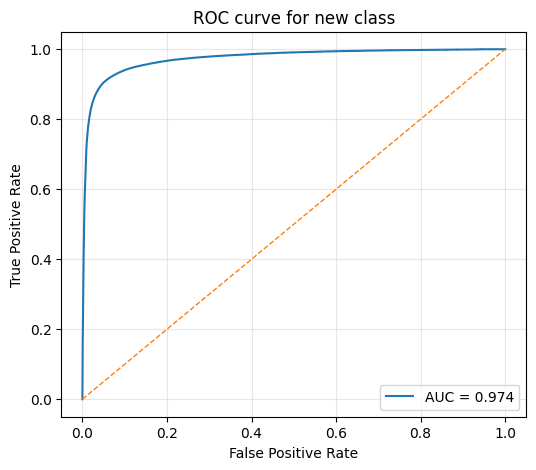} 
    \\ \vspace{0.1cm} (b)
    
    \caption{(a) Evaluation of IoU, Precision, and Recall across various decision thresholds ($\tau$) for the novel class. The peak IoU is observed at $\tau = 0.75$. (b) The ROC curve for the new class prediction, demonstrating high discriminative accuracy with an Area Under the Curve (AUC) of 0.974.}
    \label{fig:threshold_roc}
\end{figure}

The performance metrics for the novel class, including Intersection over Union (IoU), Precision, and Recall, were analyzed across a range of threshold values. As illustrated in Figure \ref{fig:threshold_roc}(a), increasing the threshold naturally improves Precision by filtering out low-confidence predictions, but simultaneously degrades Recall as more true positive pixels are rejected. The optimal balance is achieved at $\tau = 0.75$, where the IoU reaches its peak at 0.5314. At this threshold, the model demonstrates a strong F1-score of 0.694, with a Precision of 0.661 and a Recall of 0.731.

Furthermore, a Receiver Operating Characteristic (ROC) analysis was conducted to evaluate the discriminative capability of the new class head across all possible thresholds. As shown in Figure \ref{fig:threshold_roc}(b), the model exhibits excellent classification performance for the incremental category, achieving an Area Under the Curve (AUC) of 0.974. Based on these quantitative findings, the optimal threshold of $\tau = 0.75$ was established and utilized for all subsequent incremental learning evaluations.

\subsection{Qualitative Analysis}
\label{sec:qualitative}
\begin{figure}[H]
\begin{adjustwidth}{-\extralength}{0cm}
\centering
\includegraphics[width=\linewidth]{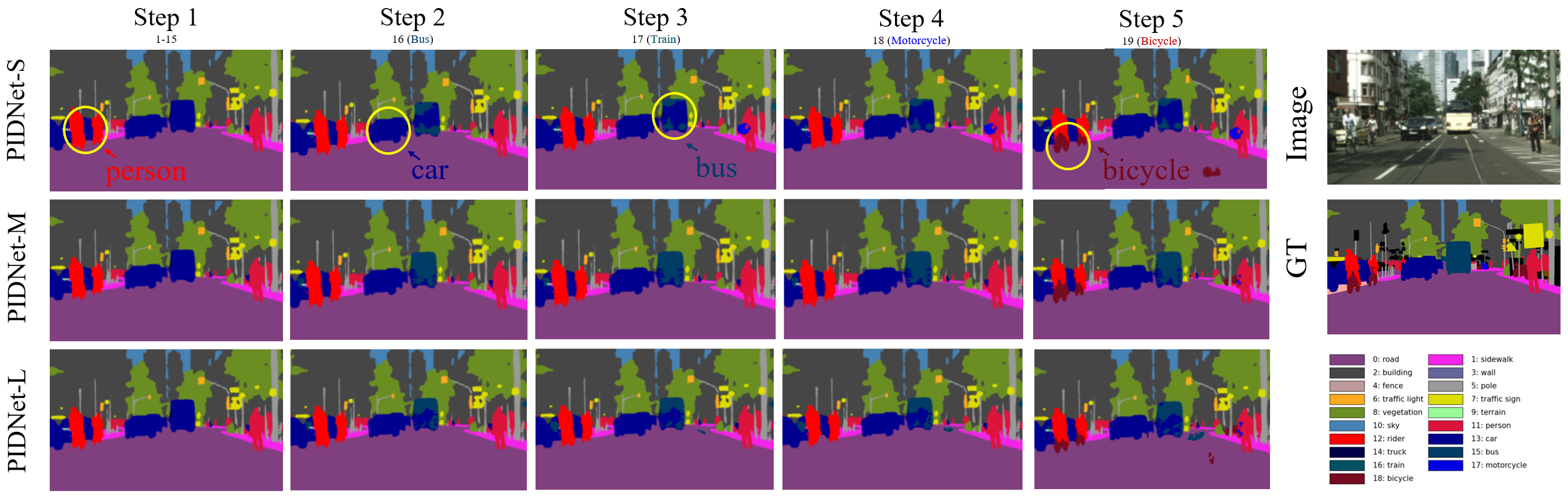} 

\includegraphics[width=\linewidth]{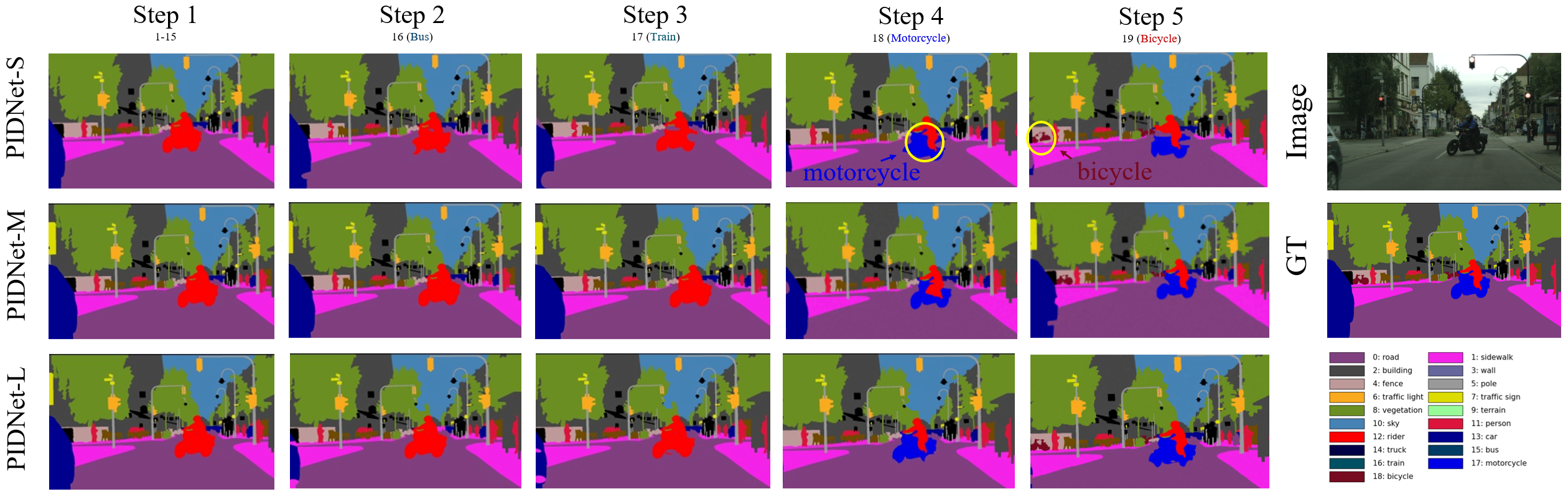}
\end{adjustwidth}
\caption{Qualitative evolution of incremental segmentation. The rows correspond to different model capacities (PIDNet-S, M, and L). The columns display the model's predictions at each incremental step: \textbf{Step 1} (Base 15 classes), \textbf{Step 2} (+Bus), \textbf{Step 3} (+Train), \textbf{Step 4} (+Motorcycle), and \textbf{Step 5} (+Bicycle). The final column shows the original Input Image and the Ground Truth (GT). Note how the models successfully segment the new "Bicycle" class (Step 5) while preserving the boundaries of the "Person" and "Car" classes learned in Step 1.\label{fig:qualitative_results}}
\end{figure}

To provide visual insight into the model's incremental learning capability, we present the qualitative segmentation results in Figure \ref{fig:qualitative_results}. We visualize the predictions across three model capacities (PIDNet-S, M, and L) as the model sequentially learns four new classes: Bus (Step 16), Train (Step 17), Motorcycle (Step 18), and Bicycle (Step 19).

As observed in the figure, our method demonstrates high plasticity throughout the incremental process. At Step 2, the model successfully identifies the ``Bus'' (highlighted in dark blue), effectively distinguishing it from the surrounding traffic. As the learning progresses to Step 5, the model accurately segments the ``Bicycle'' (dark red). This is a particularly challenging scenario, as bicycles possess thin structural elements that often visually overlap with the ``Person'' class. Despite this complexity, our method generates sharp boundaries for the new objects without eroding the segmentation quality of the previously learned categories.

Equally important is the stability of the base classes. In standard fine-tuning approaches, the introduction of new categories often leads to catastrophic forgetting, where the model begins to confuse previously learned textures with the new object as shown in Figure \ref{fig:intro_forgetting}. However, looking at the progression from Step 1 to Step 5, the segmentation masks for base classes such as ``Vegetation,'' ``Car,'' and ``Person'' remain consistent and distinct. Comparing the model variants, the PIDNet-M backbone (middle row) exhibits the most stable boundaries. While PIDNet-S misses some fine details on the bicycle and PIDNet-L introduces slight noise in the background, PIDNet-M maintains a clean separation between the new bicycle class and the existing person class, visually confirming our quantitative finding that the medium-capacity backbone offers the optimal balance for this task.

\section{Discussion} \label{sec:Discussion}

The results of this study provide critical insights into the stability plasticity trade off within real-time semantic segmentation architectures. This study proposes a novel replay free continual learning framework. It directly addresses the critical challenge of catastrophic forgetting in dynamic environments. The methodology leverages the parallel architecture of the PIDNet model. It structurally isolates the learning of novel semantic categories from previously retained knowledge. The ablation analysis explicitly demonstrates the optimal design. Attaching a lightweight unfrozen parallel branch exclusively to the high frequency outputs of the frozen Derivative (D) branch provides the best balance between plasticity and stability. This allows the model to acquire new spatial representations without corrupting the broader contextual understanding maintained by the Proportional (P) and Integral (I) branches.

Furthermore, the capacity analysis reveals an important dynamic regarding model scaling in continual learning. It is generally assumed that larger models inherently perform better. However, the findings indicate that the medium capacity backbone (PIDNet-M) consistently outperforms the larger variant (PIDNet-L) across sequential incremental steps. Oversized networks may suffer from overfitting to the limited data of the novel classes. This leads to unnecessary disruption of the frozen base weights. The medium backbone provides the optimal structural balance. It possesses sufficient capacity to learn complex new features while maintaining the rigidity required to resist catastrophic interference.

Comprehensive evaluations in a class incremental learning scenario reveal highly effective results. The PIDNet-M model successfully learned new classes without overwriting previously retained knowledge. It maintained a robust overall mIoU of 71.87\%. Most importantly, the framework preserved a high base class stability of 77.38\%. The proposed methodology achieves these metrics without relying on computationally expensive multi scale feature distillation or memory intensive data replay. Conventional approaches are fundamentally designed for offline and high capacity models like ResNet 101. This introduces significant computational overhead. By structurally isolating the learning process, the computational footprint remains minimal. This successfully bridges the gap between advanced continual learning strategies and the strict latency requirements of real-time segmentation models.

Despite the strong results, there are still several directions worth exploring in future work. First, the proposed framework is currently built on top of a convolutional dual resolution backbone, where the clear separation between the Proportional, Integral, and Derivative branches makes it natural to attach a parallel boundary branch. A logical next step is to test whether the same idea works beyond this single type of architecture. Transformer based segmentation models, and other dense prediction tasks such as object detection, also rely on a mix of contextual and high frequency features. It would be valuable to study how a frozen backbone paired with a small unfrozen branch can be applied to attention based encoders, or to detection heads that need to keep learning new object categories over time. This would help confirm whether the stability and plasticity benefits seen here come from a general design principle, rather than from something specific to PIDNet. Second, the current evaluation is mainly focused on driving datasets such as Cityscapes, where the classes and scenes are closely tied to vehicles, roads, and urban environments. Early experiments on Pascal VOC show that the method can transfer in principle, but the accuracy on more diverse classes is not yet as strong as the in domain results. Future work will therefore aim to extend the method to more general and varied segmentation datasets. This includes looking at how the boundary cues from the Derivative branch behave when new classes are non-rigid, deformable, or visually very different from the base classes, and finding ways to keep the parallel branch lightweight while still handling the wider range of objects needed for use beyond autonomous driving.

\section{Conclusion} \label{sec:Conclusion}

In conclusion, PILOT successfully bridges the gap between advanced continual learning strategies and the strict latency constraints of lightweight segmentation models. Unlike traditional frameworks that rely on heavy offline architectures, the proposed method preserves high frame rates and low inference latency. These characteristics are absolutely essential for real-time processing. It offers a highly efficient and deployment ready solution for real world autonomous driving systems. For everyday applications, this means autonomous vehicles, delivery robots, and smart city cameras can continuously learn to recognize new things on the fly, including novel vehicles and temporary road barriers. By maintaining real-time perception speeds while entirely eliminating the need for costly offline retraining, this framework ensures that autonomous systems can safely and instantly adapt to rapidly changing environments.

\vspace{6pt} 

\authorcontributions{Conceptualization, Y.Z. and P.S.; methodology, Y.Z. and P.S.; software, Y.Z.; validation, Y.Z.; formal analysis, Y.Z. and P.S.; investigation, Y.Z.; resources, P.S. and Y.L.; data curation, Y.Z.; writing---original draft preparation, Y.Z.; writing---review and editing, Y.Z., P.S., T.Y. and Y.L.; visualization, Y.Z.; supervision, P.S., T.Y. and Y.L.; project administration, Y.L.; funding acquisition, Y.L. All authors have read and agreed to the published version of the manuscript.}

\funding{This research was funded by the Cyber-CARE Center.}

\institutionalreview{Not applicable.}

\informedconsent{Not applicable.}

\dataavailability{The Cityscapes dataset used in this study is publicly available at \url{https://www.cityscapes-dataset.com/}. The source code implementing our PILOT framework is available at \url{https://github.com/U1overground/PILOT}.}

\conflictsofinterest{The authors declare no conflicts of interest.}

\begin{adjustwidth}{-\extralength}{0cm}

\reftitle{References}

\bibliography{references}

\end{adjustwidth}
\end{document}